\documentclass{article}
\usepackage[preprint, nonatbib]{neurips_2024}

\usepackage[utf8]{inputenc} 
\usepackage[T1]{fontenc}    
\usepackage[usenames,dvipsnames]{color}
\usepackage{hyperref}       
\hypersetup{
    bookmarks=false,
    colorlinks,
    linkcolor=Black,        
    citecolor=RubineRed,             
    filecolor=Plum,      	    
    urlcolor=PineGreen,         
}
\usepackage{url}            
\usepackage{booktabs}       
\usepackage{multirow, multicol}
\usepackage{amsfonts}       
\usepackage{amsmath}
\usepackage{nicefrac}       
\usepackage{microtype}      
\usepackage{bm}
\usepackage[ruled]{algorithm2e}
\SetKwComment{Comment}{/* }{ */}
\usepackage[backend=biber, style=nature, maxnames=10]{biblatex}
\usepackage{graphicx}
\usepackage{lipsum}
\usepackage{siunitx}
\title{Do Graph Neural Networks Work \\for High Entropy Alloys?}

\author{
  Hengrui Zhang\\Northwestern University\\
  \texttt{hrzhang@u.northwestern.edu}\\
  \And
  Ruishu Huang\thanks{Work performed while at Northwestern University.}\\University of Wisconsin--Madison\\
  \texttt{rhuang255@wisc.edu}\\
  \And
  Jie Chen$^*$\\Virginia Tech\\
  \texttt{jiechen@vt.edu}\\
  \And
  James M.~Rondinelli\\Northwestern University\\
  \texttt{jrondinelli@northwestern.edu}\\
  \And
  Wei Chen\\Northwestern University\\
  \texttt{weichen@northwestern.edu}\\
}

\begin{document}

\maketitle

\begin{abstract}
Graph neural networks (GNNs) have excelled in predictive modeling for both crystals and molecules, owing to the expressiveness of graph representations. High-entropy alloys (HEAs), however, lack chemical long-range order, limiting the applicability of current graph representations. To overcome this challenge, we propose a representation of HEAs as a collection of local environment (LE) graphs. Based on this representation, we introduce the LESets machine learning model, an accurate, interpretable GNN for HEA property prediction. We demonstrate the accuracy of LESets in modeling the mechanical properties of quaternary HEAs. Through analyses and interpretation, we further extract insights into the modeling and design of HEAs. In a broader sense, LESets extends the potential applicability of GNNs to disordered materials with combinatorial complexity formed by diverse constituents and their flexible configurations.
\end{abstract}


\section{Introduction}
Advances in machine learning (ML) have greatly facilitated the understanding and design of materials. In particular, graph neural networks (GNNs) have undergone extensive development and shown outstanding performance across various tasks for both crystalline and molecular materials \cite{gnnrev-comm-22}, mainly benefiting from the versatility of graph representation in capturing local chemical environments. For molecules, a graph representation is straightforwardly defined by encoding atoms as nodes and chemical bonds as edges \cite{neurmp-icml-17}. This method has then been adapted to crystals \cite{cgcnn-prl-18}, whose repeating units are represented as graphs, with atoms as nodes and nearest neighbors connected by edges. Moreover, graphs are flexible to incorporate physics information, such as electrical charge \cite{chgnet-nmi-23}, spin \cite{spingnn-prb-24}, or symmetry \cite{e3nn-icml-21}, which is highly desirable for materials science applications.

Adopting GNNs to high-entropy alloys (HEAs) has been deemed challenging \cite{mlhea-prog-23}. Single-phase HEAs are chemically disordered alloys formed by multiple metal elements often with near-equimolar concentrations that can lead to short-range correlations \cite{taheri-mrsb-23}.
They form a promising candidate space for materials with desired properties ranging from thermal \cite{mlhea-science-22}, mechanical \cite{toughhea-science-22} to catalysis \cite{heacat-jourle-19}. Owing to the lack of periodic (long-range) chemical order of the multiple species, atomistic representations of HEAs with graphs are not as easy to construct as they are for molecules or crystals. Existing ML models for HEAs mainly use tabular descriptors \cite{nnhea-matter-20, mlhea-co2-22, mlelast-acta-22} or graph representation of the composition \cite{ecnet-npj-22}, whereas local environments are not considered.

Our recent work \cite{molsets-prxe-24} proposes a ``graph set'' representation for molecular mixtures, along with an ML model architecture named MolSets to learn molecular chemistry and make predictions of mixture properties. We recognize that HEAs have similar characteristics as molecular mixtures: both are unordered collections of local chemical environments (molecules or atomic sites). With this analogy, we propose a graph set representation of HEAs and an extended ML model LESets to predict the properties of HEAs represented thereby. The remaining sections are organized as follows: Section \ref{sec:related} reviews the literature of related works. In Section \ref{sec:methods}, we describe the proposed representation and model. We then present results including benchmark, sensitivity analysis, and model interpretation in Section \ref{sec:results}. Section \ref{sec:discuss} discusses the benefits, outlook, and limitations.

\section{Related Works} \label{sec:related}
In this section, we briefly review the literature in two relevant directions: (1) GNNs for materials property prediction and (2) computational modeling of HEAs at the atomic scale.

\paragraph{GNNs for materials.}
GNNs have become an essential method in materials structure--property modeling. The key to building such models is to represent the structure of a material as a graph, a data structure with nodes and edges. Intuitively, graphs are formulated with atoms as nodes, and atoms that share electrons (bonded or are nearest neighbors) are connected by edges, with atomic/bonding descriptors as node/edge features. On graph data, GNNs perform ``message passing'' to fuse node/edge information and predict graph-level outputs, such as materials properties. Following this approach, GNNs have been applied to molecules \cite{neurmp-icml-17} and crystals \cite{cgcnn-prl-18}. Yet, the applicability is limited to materials with well-defined structures or repeating units, whereas for unordered, combinatorial materials systems, such as HEAs, a graph representation is hard to formulate.

Beyond the simple formulation, graph representation and GNNs are flexible to incorporate various physical information or knowledge. Chemprop \cite{chemprop-jcim-24} employs bi-directional message passing to attain high accuracy in molecular property prediction. ALIGNN \cite{alignn-npj-21} represents bonds and angles by nodes and edges, respectively, thus incorporating angular information in crystals. M3GNet \cite{m3gnet-ncs-22} and MACE \cite{mace-nips-22} utilize many-body interactions; CHGNet \cite{chgnet-nmi-23} and SpinGNN \cite{spingnn-prb-24} encode electrical charge and spin on atoms, respectively. Another broad category of physical information is symmetry, which is incorporated in GNNs in the form of invariance or equivariance \cite{e3nn-icml-21, painn-icml-21}. As the model proposed in this work is a generic framework agnostic of the exact form of GNN, these are orthogonal yet complementary directions that could be combined with the framework we propose.

\paragraph{Modeling HEAs.}
One key challenge to the computational modeling of HEAs is the randomness, or lack of order, in their structures. The atomic scale physics-based modeling techniques, such as density functional theory and molecular dynamics, require well-defined structures as input. Due to the lack of order in HEAs, such studies generally use (special quasi) random structures given composition, construct large supercells to account for the randomness, or construct cluster expansions \cite{mdhea-ijms-23, dfthea-jpcc-23, widom-jmr-18}.

Machine learning models take various approaches to handle or bypass such randomness. One way is to form a tabular input data structure from summary statistics of elemental descriptors and model it using conventional ML methods, such as gradient boosting and random forest \cite{gbrf-md-24}, or neural networks. Other ways include constructing input representations from multiple random realizations of structures or frequency counting \cite{dnnhea-acscat-24}. There are also GNN methods, such as ECNet \cite{ecnet-npj-22}, working on graph representations of composition alone. However, the capability of GNNs to capture connectivity and interactions in local chemical environments has not yet been unleashed in HEAs.

\section{Methods} \label{sec:methods}
\subsection{Representing HEA by local environments}
\begin{figure}[ht]
    \centering
    \includegraphics[width=0.75\linewidth]{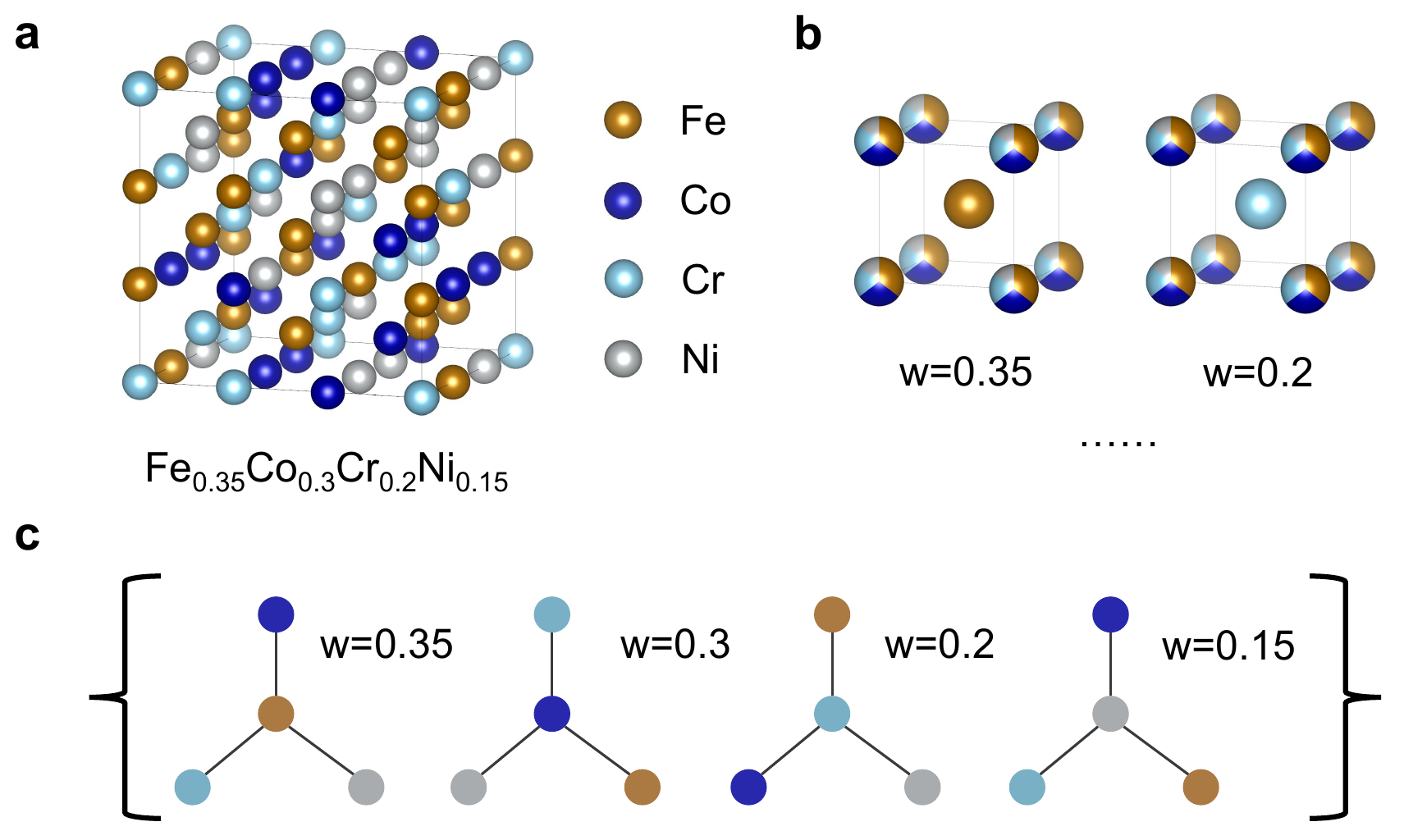}
    \caption{(a) Illustration of the atomic-scale structure of a quaternary HEA. (b) The local environments. (c) Representing the HEA as a set of local environment graphs.}
    \label{fig:represent}
\end{figure}

As \autoref{fig:represent}(a) shows, a typical HEA structure is unordered at the atomic scale. Specifically, the atomic sites in the structure may be fixed, but the arrangement (or decoration) of atoms among those sites is nearly random. To formulate a graph representation, we focus on local environments (LEs), i.e., an atom and its nearest neighbors. For an atom (e.g., Fe), its neighbors may be all types of elements, appearing at frequencies determined by their molar fractions. We represent this local environment by a graph, where the Fe atom is the center node connected to all other types of atoms, whose molar fractions are the edge weights. Note that though the Fe node is not directly connected to itself, its information could propagate to itself via message passing. Each node is associated with a feature vector of elemental descriptors: one-hot encoded elemental period and group, atomic mass (Dalton), covalent radius (\text{\AA}), Pauling electronegativity, first ionization energy (eV), electron affinity (eV), and atomic volume (\si{\cubic\centi\meter\per\mol}). The weight fraction of a non-center atom is used as an edge feature. Every type of LE is thus represented as a graph.

The HEA as a whole is an unordered collection of different local environments. For example, the quaternary HEA shown here consists of four types of LEs, centered by Fe, Co, Cr, and Ni, respectively. We represent the HEA as a set of four LE graphs $x$, each associated with a weight fraction $w$ that is the fraction of the center element in the HEA's composition, formally, $X=\{(x_i,w_i)\}$. This graph set representation captures the atomic interactions of LEs, as well as the flexibility of relations between LEs, matching the unordered configuration of LEs in the HEA structure. 

\subsection{LESets model}
In \cite{molsets-prxe-24}, we present MolSets that integrates GNN with the permutation invariant Deep Sets architecture \cite{deepstes-nips-17} and attention mechanism \cite{attention-nips-17} for molecular mixture property modeling. In MolSets, each molecule is represented as a graph, and a mixture is represented as a weighted set of graphs. Noticing the analogy of this data structure to the LE graph-based representation of HEAs, we extend MolSets to HEAs, named LESets.

\begin{figure}[ht]
    \centering
    \includegraphics[width=0.95\linewidth]{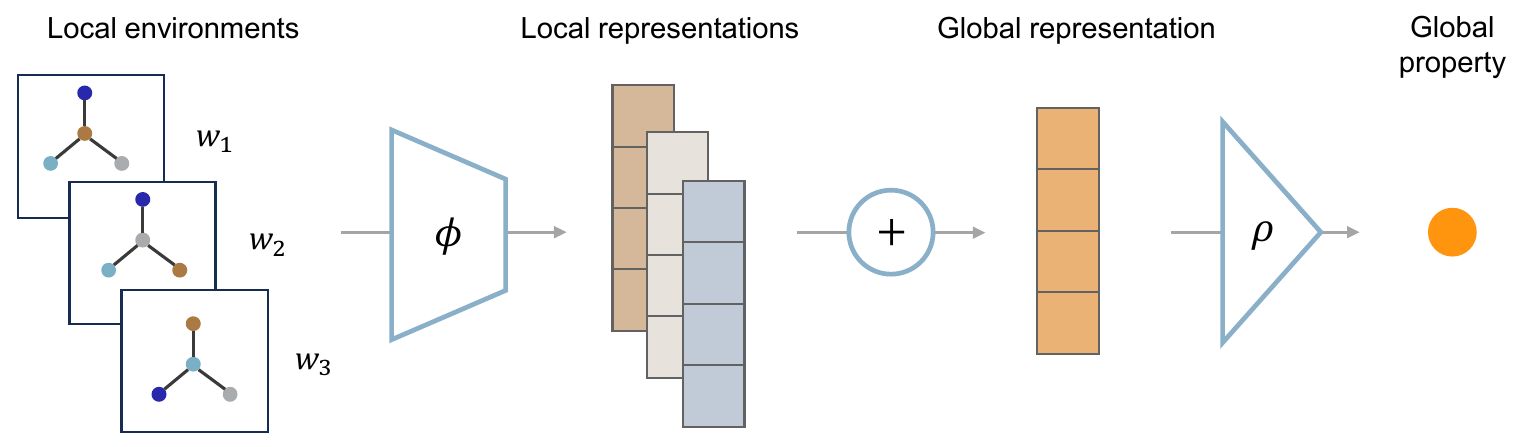}
    \caption{Architecture of LESets: A graph neural network $\phi$ learns the representations of local environments. Aggregation module $\oplus$ combines the local environment representations and their weights to form a global representation of the HEA. A multilayer perceptron $\rho$ maps the global representation to target materials property.}
    \label{fig:architecture}
\end{figure}

LESets consists of three modules, as shown in Figure \ref{fig:architecture}. Taking an input datapoint $X=\{(x_i,w_i)\}$, a GNN module $\phi$ maps each local environment graph $x_i$ to its intermediate representation vector $z_i$, i.e., $z_i=\phi(x_i)$. Then, $\{(z_i,w_i)\}$ is aggregated into a global representation vector $Z$, which is mapped to the target response $y$ (HEA property) by a multilayer perceptron $\rho$.
Overall, the model learns an input--response function of the following form:
\begin{equation}
    y = \rho \left(\oplus_{(x,w)\in X}\left\{\phi(x),w\right\}\right).
\end{equation}
Specifically, for the aggregation $\oplus$, we take two approaches: 
\begin{enumerate}
    \item Weighted summation of LE representations, i.e.,
\begin{equation}
    \oplus \{\phi(x_i), w_i\} = \sum w_i\phi(x_i);
\end{equation}
    \item Adjust LE representations using attention mechanism:
    \begin{equation}
        z_i = \texttt{attention}[\phi(x_i)]~\textrm{for every $x_i$ in $X$},
    \end{equation}
taking into account the disproportionate importance and interactions between LEs within a HEA, before weighted summation $\oplus{\{z_i, w_i\} = \sum w_iz_i}$.
\end{enumerate}

We name the model using each approach ``LESets-att'' and ``LESets-ws'', respectively. A formal description of the model is provided in Algorithm \ref{algo:lesets} within Appendix \ref{sec:algo}.

\section{Results} \label{sec:results}
\subsection{Data and experimental setup}
To test and demonstrate the performance of the LESets model, we conduct experiments on a previously reported DFT-calculated HEA property dataset \cite{deepsets-hea-22}. The dataset contains bulk modulus $B$ and Wigner-Seitz radius $r_{\rm ws}$ for 7086 HEAs, 1909 of which also have Young's modulus $E$ reported. We test our and other ML models in predicting these properties and measure their performances using two metrics, coefficient of determination $R^2$ and mean absolute error (MAE).

We compare the LESets model with several previous methods: (1) The Deep Sets model with HEAs represented as sets of elements, as presented in \cite{deepsets-hea-22}. Note that the major distinction of LESets from this model is incorporating local environment information encoded in graphs. (2) Conventional statistical learning methods, including gradient boosting decision trees, random forest, k-nearest neighbors (kNN), support vector machine (SVM), Lasso, and ridge regression, using elemental property descriptors, all implemented in \texttt{sklearn} \cite{sklearn-jmlr}. For LESets, we also investigate the effect of using attention in the $\oplus$ module.

Experiments of the deep learning models (Deep Sets and LESets) are conducted with the following procedure. First, the hyperparameters of every model and target property are tuned using grid search on one random split of the dataset. The optimal hyperparameters found by tuning are listed in Table \ref{tab:hyperpar} within Appendix \ref{sec:supp}. Then, the dataset undergoes 30 random splits into training, validation, and test data in a 3:1:1 ratio. In each split, the model is trained by minimizing mean squared error (MSE) loss on the training dataset using the AdamW optimizer \cite{adamw}, with a weight decay coefficient of $10^{-4}$ for regularization. The model's loss on the validation dataset is monitored for learning rate scheduling (learning rate is reduced by half after 10 epochs of no improvement) and early stopping (training is terminated after 20 epochs of no improvement). After training, the model makes predictions on the testing dataset, and its performance metrics are recorded. The models are implemented in PyTorch and PyTorch Geometric (PyG). 

For the conventional ML models, experiments are conducted similarly: the hyperparameters are tuned using grid search and 5-fold cross-validation on 40\% of the dataset, then their performances are assessed on 30 random splits of the remaining data into training and testing in a 2:1 ratio. Computational details are provided in Appendix \ref{sec:algo}.

\subsection{Benchmark test results}
\begin{figure}[ht]
    \centering
    \includegraphics[width=0.8\linewidth]{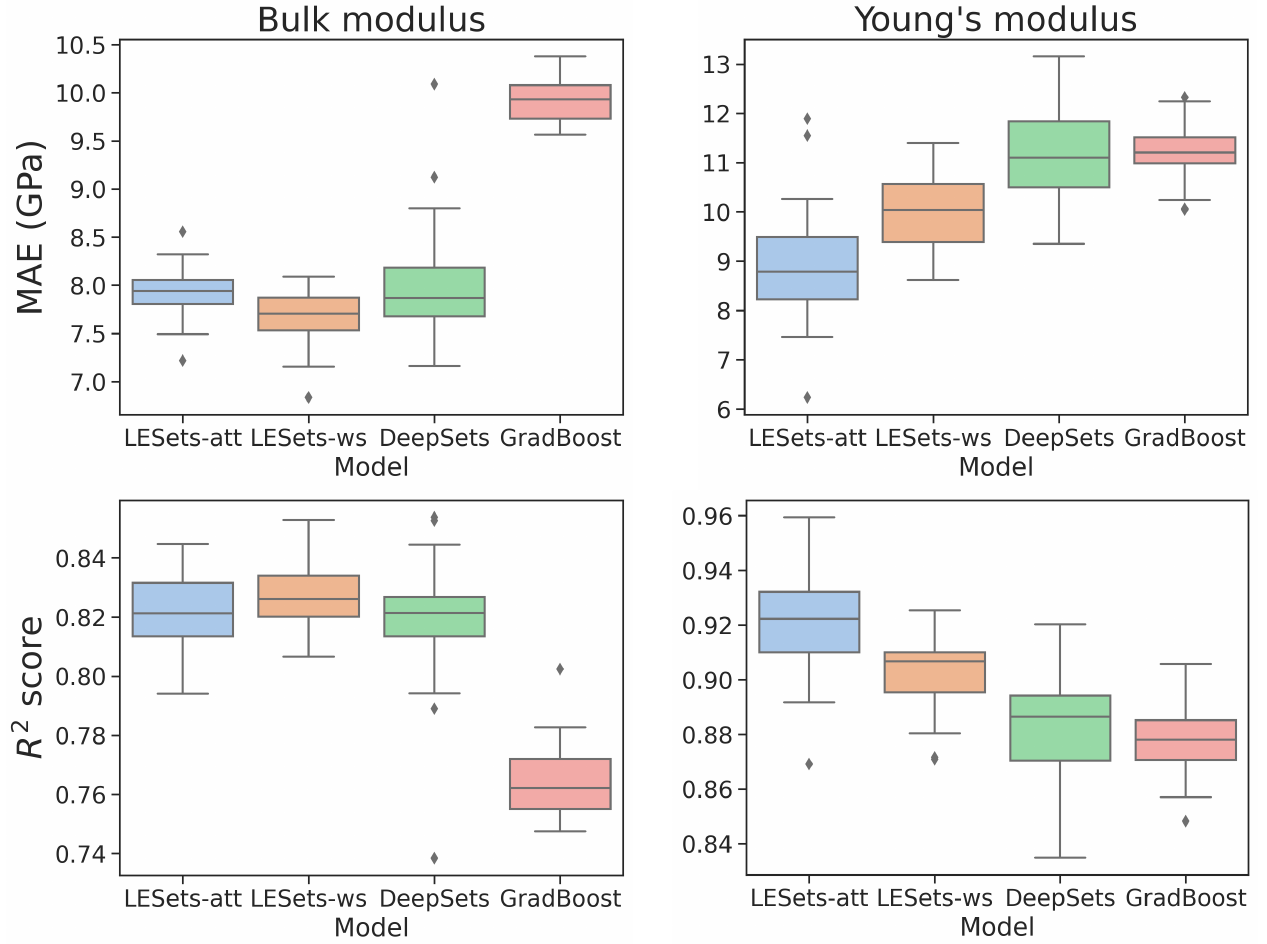}
    \caption{MAE and $R^2$ of LESets, Deep Sets, and gradient boosting on the testing dataset across 30 random data splits.}
    \label{fig:boxplots}
\end{figure}

\begin{figure}[ht]
    \centering
    \includegraphics[width=0.95\linewidth]{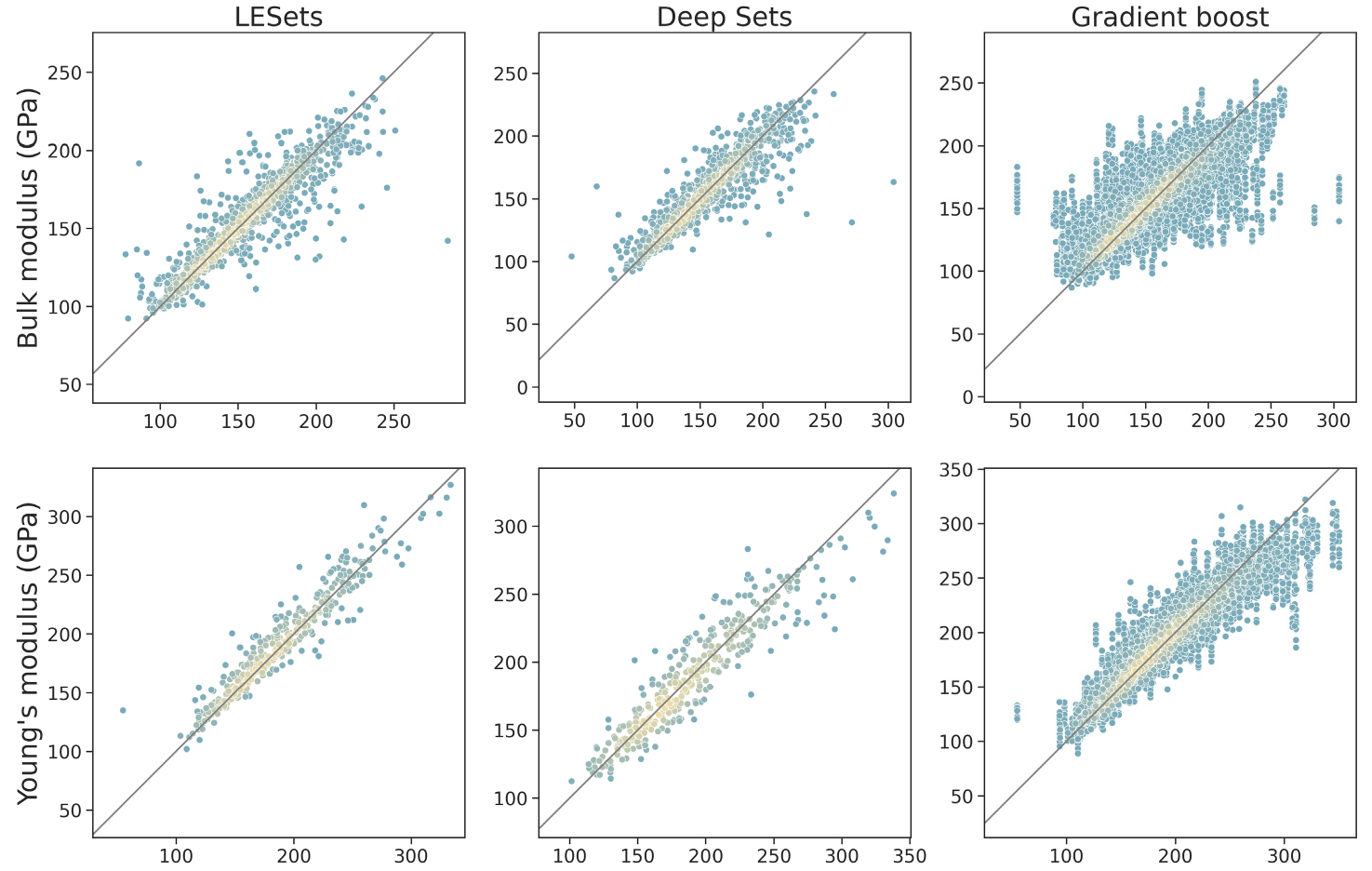}
    \caption{Regression plots showing target values (horizontal axis) and predicted values (vertical axis) of $E$ and $B$ for LESets (the better one between LESets-ws and LESets-att), Deep Sets, and gradient boosting. Colors reflect the density of points, quantified by kernel density estimation.}
    \label{fig:regplots}
\end{figure}

In Figure \ref{fig:boxplots}, we show the performance metrics on predicting bulk modulus $B$ and Young's modulus $E$ for LESets, Deep Sets, and gradient boosting, which is the best among conventional ML models. As all models attain high $R^2$ and low error for $r_{\rm ws}$, we exclude it from comparison here. We list the average performance metrics of all tested models and properties in Table \ref{tab:ml-compare} within Appendix \ref{sec:supp}. To visualize different models' predictions, we show their predicted values vs.~true values in one of the random data splits in Figure \ref{fig:regplots}.

From the results, we find that LESets attains higher prediction accuracy than its counterparts. Notably, compared to Deep Sets, $R^2$ and MAE of LESets vary less across random data splits, which indicates better robustness to data variations. The improvement in accuracy could be because local chemical environment information, such as bonding, plays an important role in determining elastic properties \cite{mech-book-08}.

A more interesting comparison is between two versions of LESets with different aggregation mechanisms. The simpler LESets-ws outperforms LESets-att in predicting $B$, suggesting that attention-based aggregation is unnecessary in this task. Whereas in predicting $E$, LESets-att regains the advantage. This difference indicates that elements or local environments may have disproportionate importance in determining $E$ but not for $B$. Further investigations using electronic structure methods could find physical insights and guide the design principles of HEAs' elastic properties.

\subsection{Sensitivity analysis}
As a deep learning model, LESets requires sufficient data to be trained. Although high-throughput computation and data platforms have made materials data more available than before, understanding the data requirement guides model usage as well as data collection efforts \cite{redundancy-nc-23}. To this end, we conduct sensitivity analyses of LESets' performance with respect to data size.

\begin{figure}[ht]
    \centering
    \includegraphics[width=0.8\linewidth]{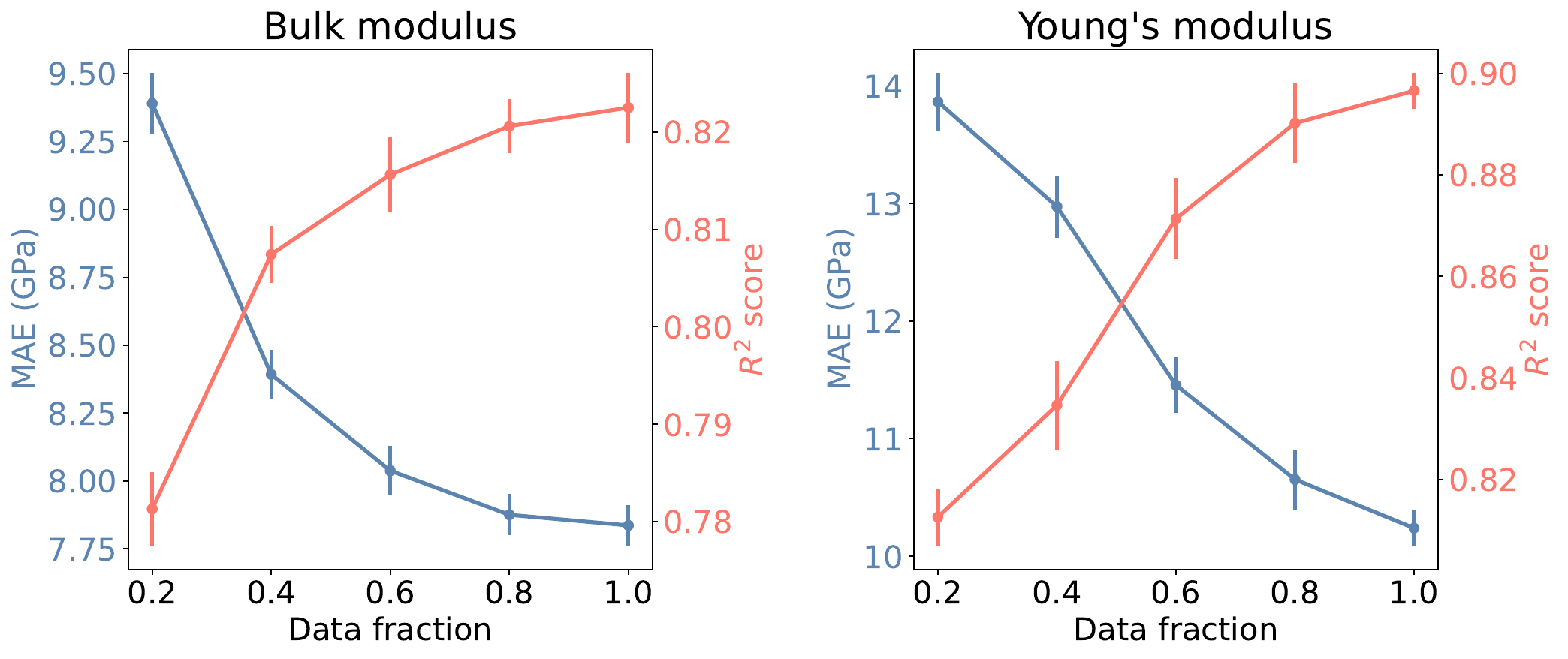}
    \caption{The change of LESets' MAE and $R^2$ on two target properties with data fraction. Points denote mean values across 10 replicates, and error bars denote standard error (standard deviation/$\sqrt{10}$).}
    \label{fig:sensitivity}
\end{figure}

The analyses are conducted on the performant version of LESets on bulk and Young's moduli, respectively, viz.~LESets-ws for $B$ and LESets-att for $E$. For each property, we leave out 20\% of the dataset to be the testing dataset and use a fraction of the remaining data in model training. This fraction of data is split into training and validation datasets in a 3:1 ratio. We repeat this experiment 10 times with different random states for every fraction. Figure \ref{fig:sensitivity} shows the relation between performance metrics and data fraction. For $B$, LESets' performance begins saturating when the data fraction is high, whereas for $E$, it maintains a trend of increasing within the range of analysis. This can be partly explained by the much larger number of datapoints for $B$ (7086) than that for $E$ (1909). Note that LESets' number of trainable parameters typically ranges from 5000--8000 (varies depending on hyperparameters). The sensitivity analysis results suggest that the number of parameters might be a rough guideline for the required amount of data.

\subsection{Interpretation}
\begin{figure}[ht]
    \centering
    \includegraphics[width=\linewidth]{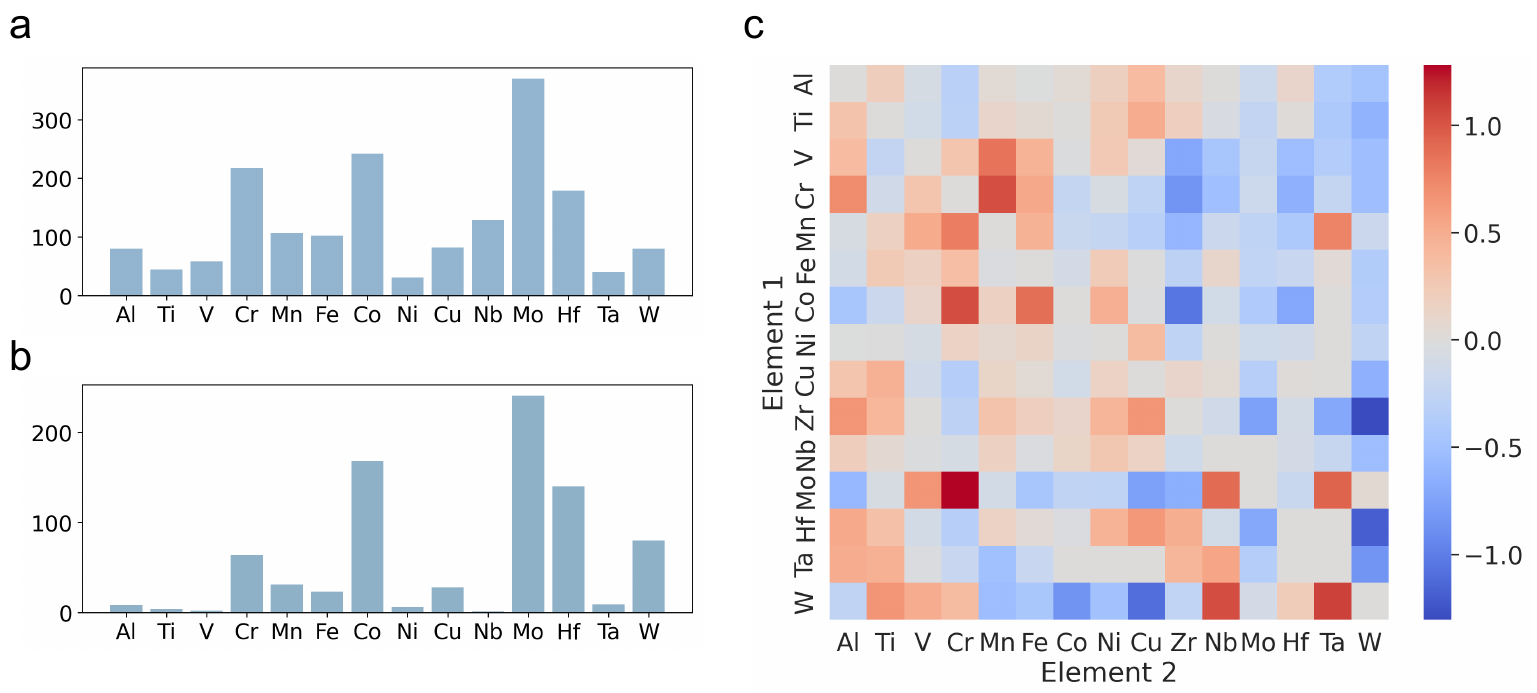}
    \caption{Frequency of (a) an element's $Imp$ being at least $3\times$ the least important one, and (b) elements being the most important in a HEA while satisfying the previous criterion. Zr is excluded because of zero appearance. (c) Difference between Element 1's average $Imp$ in all HEAs with and without Element 2.}
    \label{fig:interpret}
\end{figure}

One benefit of LESets-att is its interpretability provided by the attention-based aggregation. Following the approach in \cite{molsets-prxe-24}, we investigate the Euclidean norm change of a local environment's representation vector after attention:
\begin{equation}
    Imp = \left\lVert z'\right\rVert_2 /  \left\lVert z\right\rVert_2
\end{equation}
as a score that reflects its relative importance in the HEA. For simplicity, we refer to the importance of a local environment as the importance of its center element. With this importance score, we use two criteria to find out which elements contribute more to a HEA's Young's modulus $E$: (1) elements whose importance score is at least three times the lowest in the HEA; (2) elements with the largest score in the HEA. We plot the frequency of elements satisfying criterion (1) in Figure \ref{fig:interpret}(a), and of those satisfying both criteria in Figure \ref{fig:interpret}(b). The identified elements, such as Cr, Co, Mo, Hf, and W, appear to be frequent choices in previous studies designing high-stiffness HEAs \cite{mechhea-book-21}.

Besides the importance of single elements, we are also interested in the interactions between elements. In Figure \ref{fig:interpret} (c), we show how an element's importance score is affected by the presence of another element. Though not identified as ``important'' by its own $Imp$, elements such as Mn, Nb, and Ta strongly correlate with other elements' contributions to $E$, and are thus beneficial as additives. Theories behind the additives' effects, such as mixing enthalpy, have been explored \cite{mixing-nature-24} and remain an active research topic. We envision these and further interpretations enabled by LESets can help derive useful design principles for HEAs and guide materials discovery.

\section{Discussions} \label{sec:discuss}
The methodological advances of LESets are two-fold. Firstly, the graph set representation effectively captures the chemistry of the local environment while maintaining flexibility to cope with the lack of long-range orders in HEAs. Secondly, the use of attention mechanism in aggregation enables finding disproportionate contributions of certain ingredients (LEs in this case). It also allows a qualitative assessment of the ingredients' importance. This interpretability is useful for guiding materials design and is desired for scientific ML models.

Note that the focus of this work is to highlight the applicability of graph representation and GNNs to HEAs, while accuracy is less emphasized. Hence, the accuracy of LESets shown in Section \ref{sec:results} potentially has room for improvement. Here, we suggest two possible approaches to further optimizing its accuracy: (1) Revisit the selection of elemental descriptors for constructing node features; (2) Fine-tune the layers' numbers and hidden dimensions, with a focus on trainable parameter numbers in each module of LESets.

The LESets model architecture has demonstrated general applicability for molecular mixtures in \cite{molsets-prxe-24} and HEAs in this work. These are both instances of \emph{combinatorial} materials systems, whose diverse constituents (LEs, molecules, etc.) and various configurations of constituents form an expansive space with multilevel complexity. Many other materials families, such as heteroanionic materials \cite{heteroan-am-19}, heterostructures \cite{superlattice-acs-23} and defected functional materials \cite{delibrate-matter-19}, are combinatorial materials systems or possess combinatorial complexity. LESets can potentially provide a generic and extensible framework for ML method development for these materials systems.

\paragraph{Limitations.}
At last, we discuss the limitations of LESets and future work directions toward overcoming them. (1) LESets' representation of local environment information is based on an assumption that all elements could appear in an atom's nearest neighbors, and simplified by modeling their appearance frequencies rather than exact locations. With emerging materials characterization techniques \cite{atomic-am-24}, the atomic-scale structures of HEAs can be obtained. When more exact structural data become available, LESets could utilize them to represent HEAs in a more informative way. (2) In addition to LEs and their configurations at the atomic scale, microstructure also impacts the mechanical properties of HEAs \cite{hea-micro-22}. LESets currently does not take into account microstructure. A future work direction toward more accurate modeling of HEAs' mechanical properties is extending the representation and model to incorporate multiscale information.

\section{Conclusion}
In this work, we present a representation of high-entropy alloys (HEAs) as a collection of local environments, and the LESets machine learning model to accurately predict HEAs' properties using this representation. Through benchmark tests, we demonstrate the advantage of LESets in predicting HEA mechanical properties over existing ML models. Furthermore, we conduct sensitivity analyses to provide guidelines for LESets' data requirements. Finally, utilizing LESets' interpretability, we investigate elemental contributions to Young's modulus of HEAs, deriving insights to guide HEA design.
The code will be available at \href{https://github.com/Henrium/LESets}{https://github.com/Henrium/LESets}.

\begin{ack}
This work was supported in part by the U.S.~National Science Foundation (Awards DMR-2324173 and DMR-2219489). H.Z.~was also supported by the Ryan Graduate Fellowship. J.M.R.~acknowledges support from the U.S.~Department of Navy, Office of Naval Research (Award N000014-16-12280). The authors thank Wangshu Zheng, Xiao-Yan Li, and Pengfei Ou for their valuable insights.
\end{ack}

\printbibliography

\newpage
\appendix
\section{Algorithm and computing details} \label{sec:algo}
\begin{algorithm}
\DontPrintSemicolon
\caption{LESets machine learning model.} \label{algo:lesets}
   \KwIn{$X=\{(x_i,w_i)\}$; LE graphs $x_i=(V,E)$, with nodes $V=\{v_m\}$ and edges $E=\{e_{mn}\}$}
   \KwOut{Property prediction $y=f(X)$}
   \;
   (1) $\phi$ module: Message-passing graph neural network. \;
   \For{$x_i\in X$}{
     \For{$l=1, \dots,$ \rm \texttt{n\_conv\_layers}}{
       $v_m \gets \texttt{message\_passing} (v_m, v_n, e_{mn})$ for $v_m\in V$ and $n\in \mathrm{neighbors}(m)$ \;
       $v_m \gets \texttt{activation}(v_m)$ \;
     }
     Readout: $z_i \gets \texttt{mean\_pooling} (v_m\in V)$ \;
     FC layer: $z_i \gets \tanh(Wz_i+b)$ \Comment*[r]{individual representations}
   }\;
    
   (2) $\oplus$ module: Weighted sum and (optionally) attention mechanism. \;
  \For{\rm all $z_i$}{
   \If{\rm \texttt{use\_att}}
    {
    $q \gets W^qz_i$, $k \gets W^kz_i$, $v \gets W^vz_i$\;
    $z_i \gets \texttt{softmax}\left(q\cdot k^{\rm T}/\sqrt{d_k}\right)\cdot v$\;
    }
    }
    $Z \gets \sum_i w_iz_i$ \Comment*[r]{aggregated representation}\;
    
   (3) $\rho$ module: $L$-layer fully connected (FC) neural network. \;
   \For{$l=1, \dots, L-1$}{
    $Z \gets \texttt{activation}(W^lZ+b^l)$\;
   }
   $y \gets W^LZ+b^L$ \;
   
   \Return $y$ \Comment*[r]{value predicted}
    
\end{algorithm}

\paragraph{Computing details.}
Experiments are performed on a Linux workstation with Intel Xeon W-2295 CPU (18 cores, 3.0 GHz), NVIDIA Quadro RTX 5000 GPU (16 GB memory), and 256 GB of RAM. Training of LESets typically takes 10--20 minutes on one GPU, with variations depending on data size, hyperparameters, and epochs before termination.

\section{Supplemental results} \label{sec:supp}
\begin{figure}[ht]
    \centering
    \includegraphics[width=0.5\linewidth]{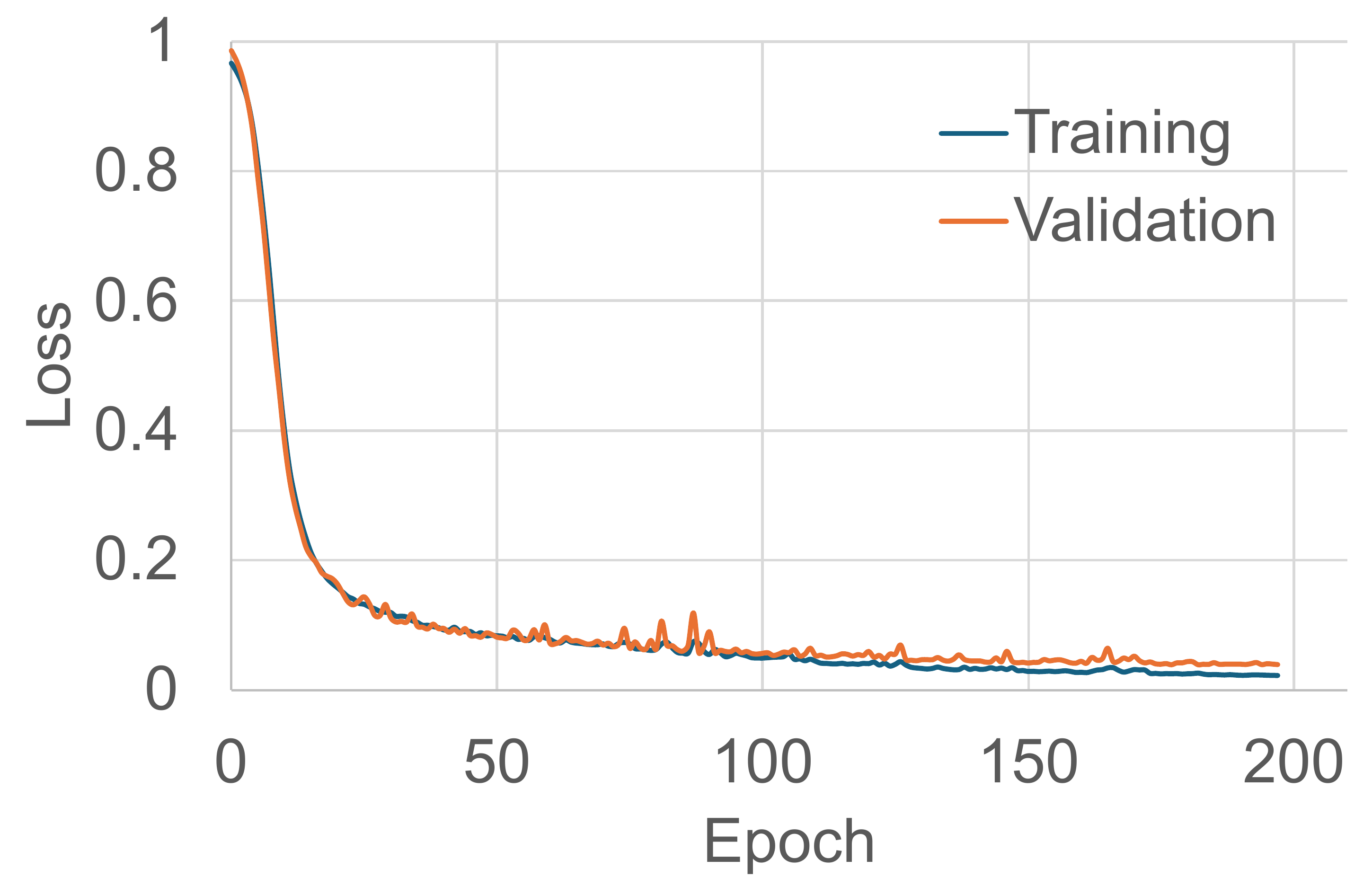}
    \caption{A typical learning curve of the LESets model.}
    \label{fig:learn-curve}
\end{figure}

Figure \ref{fig:learn-curve} shows a typical learning curve of LESets.

Hyperparameter tuning and experiment tracking are performed using the weights and biases platform \cite{wandb}; the tuning results are listed in Table \ref{tab:hyperpar}.
\begin{table}[ht]
    \centering
    \begin{tabular}{l|ccc}
    \toprule
         Hyperparameter&  Young's &Bulk &$r_{\rm ws}$\\
    \midrule
 Convolution operator & GraphConv \cite{graphconv-aaai-19}& CGConv \cite{cgcnn-prl-18} &CGConv\\
         Convolution layers& 2&3 &2\\
 FC layers& 3&3 &3\\
 Hidden dimension& 32& 32 &32\\
 Use attention & True& False &False\\
 \bottomrule
\end{tabular}
    \caption{Tuned hyperparameters of LESets for each target property.}
    \label{tab:hyperpar}
\end{table}

Table \ref{tab:ml-compare} lists the complete results of model comparisons.
\begin{table}[ht]
\centering
\begin{tabular}{lcccccc}
\toprule
\multirow{2}{*}{Model} & \multicolumn{2}{c}{Young's (GPa)} & \multicolumn{2}{c}{Bulk (GPa)} & \multicolumn{2}{c}{$r_{\rm ws}$ (\text{\AA})} \\
\cline{2-3} \cline{4-5} \cline{6-7}
                       &     MAE      &              $R^2$&      MAE    &            $R^2$&            MAE&    $R^2$\\
\midrule
                       LESets-att&              \textbf{8.891}&              \textbf{0.920}&             7.922&            0.821&            0.011&            0.989\\
                       LESets-ws&              10.03&              0.902&             \textbf{7.685}&            \textbf{0.828}&            \textbf{0.010}&           \textbf{0.989}\\
 Deep Sets \cite{deepsets-hea-22}& 11.19& 0.884& 7.984& 0.819& 0.011&0.989\\
 \midrule
 gradient boost& 11.18& 0.878& 9.935& 0.764& 0.015&0.978\\
 random forest& 14.91& 0.813& 10.87& 0.725& 0.019&0.966\\
 Lasso& 13.65& 0.838& 10.75& 0.742& 0.022&0.957\\
 ridge regression& 13.62& 0.839& 10.75& 0.741& 0.021&0.959\\
 kNN& 15.89& 0.793& 11.66& 0.728& 0.022&0.951\\
 SVM& 23.14& 0.556& 12.01& 0.662& 0.013&0.984\\
\bottomrule

\end{tabular}
\caption{Average MAE and $R^2$ across 30 replicates of all tested models on Young's modulus, bulk modulus, and Wigner-Seitz radius $r_{\rm ws}$. The best performance metric on each property is indicated by boldface.}
\label{tab:ml-compare}
\end{table}

\end{document}